\documentclass[runningheads]{llncs}
\usepackage{graphicx}

\usepackage{booktabs}       
\usepackage{amsfonts}       
\usepackage{nicefrac}       
\usepackage{microtype}      
\usepackage[table,xcdraw]{xcolor}
\usepackage{fontawesome}
\usepackage{multicol}
\usepackage{amsmath}
\usepackage{subfigure}
\usepackage{float}
\usepackage{listings}
\usepackage{arydshln}

\begin{document}
\title{Complexity-Based Code Embeddings}
\author{Rares Folea\orcidID{0000-0002-4936-9082} \and
Radu Iacob\orcidID{0009-0009-4953-1180} \and
Emil Slusanschi\orcidID{0000-0003-0222-1002} \and Traian Rebedea\orcidID{0000-0002-7255-5537}}
\authorrunning{R. Folea et al.}
%
\institute{University Politehnica of Bucharest}
\maketitle              
\begin{abstract}
This paper presents a generic method for transforming the source code of various algorithms to numerical embeddings, by dynamically analysing the behaviour of computer programs against different inputs and by tailoring multiple generic complexity functions for the analysed metrics. The used algorithms embeddings are based on r-Complexity~\cite{folea2021new}. Using the proposed code embeddings, we present an implementation of the XGBoost algorithm that achieves an average $90\%$ F1-score on a multi-label dataset with $11$ classes, built using real-world code snippets submitted for programming competitions on the Codeforces platform.

\keywords{algorithm classification, code embeddings, complexity, tree-based classification}
\end{abstract}

\section{Introduction}

\textbf{Algorithmic classification} is an important problem in Computer Science, which aims to identify the programming techniques and specific algorithms referenced in a code snippet. Solving this task requires a deep understanding of code semantics which may provide insights for many further applications, such as the detection of code vulnerabilities or the design of automatic code generation assistants.

While most of the recent community work is focused on the static code analysis, we focus on \textbf{dynamic} analysis\footnote{We define dynamic analysis as the process of developing a computer software evaluation based on data acquired from experiments conducted out on a real computing system by executing programs against a range of different inputs.}, by analysing the actual execution part to assert how the algorithm behaves. Thus, we aim to investigate how the resource usage (e.g. CPU time, memory) evolves in relation to the input size. We hypothesize that these metrics, along with more advanced architectural aspects (e.g. branch and cache misses), can be used to distinguish between different types of algorithmic approaches.

In our study, we have investigated the applicability of our approach for classifying solutions to competitive programming challenges, written in C++. Thus, we report encouraging results using decision tree models, random forest classifiers and XGBoost~\cite{chen2016xgboost} for deriving algorithmic labels from dynamic code embeddings, generated using our method.

This paper is structured as follows. In Section \ref{related-work} we report on related approaches from existing literature. Afterwards, in Section \ref{embedding-design} we describe our approach for creating dynamic code embeddings. In Section \ref{system-architecture} we present a general overview of the system architecture. In Section \ref{dataset} we provide experimental details behind the creation of a novel comprehensive dataset that we used for evaluation. Next, in Section \ref{results} we present the results achieved using several machine learning solutions. Finally, in Section \ref{conclusions} we conclude with a final perspective on the applicability of our solution.

\section{Related Work}\label{related-work}

A critical step before applying a machine learning model to analyze data is how to choose a meaningful input representation. In this section we explore several representative approaches for creating distributed program embeddings.

Firstly, researchers have investigated techniques borrowed from the field of natural language processing, which rely on the distributional properties of individual words (e.g. Word2Vec \cite{mikolov-word2vec-2018-advances}. These approaches seek to capture the statistical co-occurrence patterns of different lexical features. These features may be derived using different segmentation strategies from the textual representation of source code \cite{DBLP:journals/corr/abs-2006-12641,svyatkovskiy2021fast,DBLP:journals/corr/abs-2006-12641} or from the representation pertaining to different compilation stages (e.g. LLVM-IR \cite{neuralcodecomprehension}, Java bytecode \cite{koc2017learning}, assembly \cite{DBLP:journals/corr/abs-1812-09652}) or bytecode \cite{yousefi2018learning}. 

Another option is to train neural architectures to encode graph representations of code, such as the AST \cite{alon2019code2vec} or the CFG \cite{neuralcodecomprehension}. The advantage of this approach is that it can explicitly model semantic relationships between distant program elements.

An alternative perspective can be attained from dynamic analysis of the program runtime behaviour. One such approach is to monitor execution traces, which capture the state of variables during different moments of execution \cite{DBLP:conf/iclr/WangSS18,dypro,Liger2020}. A different direction is to collect and analyze the interactions with the operation system, such as system calls to create, open or modify files \cite{DBLP:journals/corr/abs-1804-03635}. Notably, this approach may not be suitable to distinguish between programs which perform minimal interactions with the operating system.

Static code representations entail direct access to the program source code. On the other hand, dynamic code representations discussed above leverage specific knowledge about target program semantics and rely heavily on the number and diversity of program executions. In contrast, the solution discussed in this paper requires no prior assumptions about the program structure. As such, it relies only on collected statistics about runtime behaviour, as seen through the lens of a profiling tool.


\section{Converting an algorithm to an embedding}\label{embedding-design}

As shown in \cite{folea2021new}, r-Complexity is a revised complexity model, that offers better complexity feedback for similar programs than the classic Bachmann-Landau notations. Let $f:\mathbb{N}\longrightarrow\mathbb{R}$ denote the function describing an algorithm computational complexity. We define the set of all complexity calculus $\mathcal{F}= \lbrace f:\mathbb{N}\longrightarrow\mathbb{R} \rbrace$
Also, we will consider an arbitrary complexity function $g\in \mathcal{F}$.

The \textbf{Big \textit{r-}Theta} class is defined as a set of mathematical functions similar in magnitude with $g(n)$ in the study of asymptotic behavior. A set-based description of this group can be expressed as:
\[\begin{split}
      \Theta_{r}(g(n)) = \lbrace f \in \mathcal{F}\ |\ \forall c_{1}, c_{2} \in \mathbb{R}^{*}_{+} \ s.t. c_{1}< r < c_{2} , \exists n_{0} \in \mathbb{N}^{*}\ \\ s.t.\ \ c_{1} \cdot g(n) \leq f(n) \leq c_{2} \cdot g(n)\ ,\  \forall n \geq n_{0} \rbrace
\end{split} \]

The big \textbf{r-Theta} notation proves to be useful also in creating Dynamic code embeddings. The idea behind these embeddings is simple: try to automatically provide estimations, for various metrics, the r-Theta class for the analyzed algorithm (usually with unknown Bachmann–Landau Complexity). 

A generic solution (generalization of ~\cite{calotoiu2018automatic}) to provide a good automatic estimation of the r-Complexity class is to try fit a regressor described by:

\[ f(n) =\sum\limits_{t=1}^{y}  \sum\limits_{k=1}^{x} c_{k} \cdot n^{p_{k}} \cdot log_{l_{k}}^{j_{k}}(n) \cdot e_{t}^{n} \cdot   \]

In this research, we will attempt to fit a simplified version of the generic expression as a Big r-Theta function, described generic by one of the following function:

\begin{equation*}
  \left\{\begin{array}{@{}l@{}}
    r \cdot log_{2}^{p}log_{2}(n) + X\\
    r \cdot log_{2}^{p}(n) + X \\
    r \cdot p^{n} + X, p < 1 \\
    r \cdot n^{p} + X \\
    r \cdot p^{n} + X, p > 1 \\
    r \cdot \Gamma(n) + X
  \end{array}\right.\,.
\end{equation*}

The search space for regressor functions is not feasible to be exhaustively searched during for automatic computation, as there are infinitely many regressors that can be analyzed. To overcome this problem, our approach is to avoid performing the continuous space search, but obtain similar results only by sampling a number of highly relevant configurations, that are relevant for algorithms in general. This way, we can discretize the search space, and the functions we decided to search towards are:

\begin{equation*}
  \left\{\begin{array}{@{}l@{}}
    r \cdot log_{2}^{p}log_{2}(n) + X, p \in \{0,1,2,3\}\\
    r \cdot log_{2}^{p}(n) + X, p \in \{0,1,2,...,10\} \\
    r \cdot p^{n} + X, p < 1, p \in \{0.1,0.2,...,0.9\} \\
    r \cdot n^{p} + X, p \in \{1, 1.3, 1.5, 1.7, 2, 2.5, 2.7, 3, 3.5, 4, 4.5, 5, 5.5, 6, 7, 8, 9, 10\} \\
    r \cdot p^{n} + X, p > 1, p \in \{1.5, 2, 2.5, 3, 3.5, 4, 5\} \\
    r \cdot \Gamma(n) + X
  \end{array}\right.\,.
\end{equation*}

Each fitting of a Big r-Theta on a metric has as result a quadruple: 
\begin{verbatim}
(FEATURE_TYPE, FEATURE_CONFIG, INTERCEPT, R-VAL)
\end{verbatim}

where: 

\begin{enumerate}
    \item \verb|FEATURE_TYPE| has one of the following values\footnote{The model is generic and other values can be used as well, yet these are the most relevant values that we have used in our research.}: 
        \begin{enumerate}
            \item \verb|LOGLOG_POLYNOMIAL|, 
            \item \verb|LOG_POLYNOMIAL|, 
            \item \verb|FRACTIONAL_POWER|, 
            \item \verb|POLYNOMIAL|, 
            \item \verb|POWER|, 
            \item \verb|FACTORIAL|, 
        \end{enumerate}
    \item \verb|FEATURE_CONFIG| is defined\footnote{In our research, we have searched only a small discrete set of values for $n$, described earlier in this section.} by a value $n \in \mathbb{R}_{+}$, such that the generic Big r-Theta respects the above \verb|FEATURE_TYPE|.
    \item \verb|INTERCEPT| is the expected value of the complexity function when the input size is null.
    \item \verb|R-VAL| is the value of $r$, such that the defined $g$ complexity class includes $f$, which is the real complexity function of the algorithm:
    $ f \in \Theta_{r}(g(n)) $
\end{enumerate}

Even if the model is generic and works for any arbitrary metric, in order to use it, we are required to instantiate the model with tangible metrics, with respect to the input size. We analyze an algorithm by 
all the associated r-Complexities, bundled as one code-embedding. Using the native Linux \verb|perf| profiler, we obtained and used the following metrics to compute our code embeddings. The set of these metrics covers both hardware events:
\textbf{\textit{branch-misses}} (number of branch missed predictions), \textbf{\textit{branches}} (number of branches instructions executed), \textbf{\textit{cycles}} (number of executed cycles), \textbf{\textit{instructions}} (number of executed instruction), \textbf{\textit{stalled-cycles-frontend}} (number of "wasted" CPU cycles where the frontend\footnote{Frontend refers here to the part of the hardware responsible for fetching and decoding instructions.} did not feed the backend with micro-operations.), as well as software events, \textbf{\textit{context-switches}} (number of procedures of storing the state of the process/thread, at suspension time), \textbf{\textit{CPU-migrations}} (number of times a process/thread has been scheduled on a different CPU), \textbf{\textit{page-faults}} (number of events when the virtual memory was not mapped to the physical memory), \textbf{\textit{task-clock}} (stores the clock count specific to the task that ran).  

In computing the embedding, we require no knowledge about the structure of the analyzed algorithm.
However, the resulting embedding is dependent on the architecture it has been generated against. Typically, metrics vary from architecture to architecture. The general complexity class of the algorithm may not change, thus making the \verb|FEATURE_TYPE| and \verb|FEATURE_CONFIG| parameters likely to remain the same on multiple computing architectures. The sample embeddings in this paper have been obtained when running on a 3rd Generation Intel Core processor (3.10 GHz), i5-3210M (2.5 GHz, 3MB L3 cache, 2 cores). On the other hand, given another algorithm, with a different complexity, the previously mentioned parameters are susceptible to changes.

\section{System Architecture} \label{system-architecture}

\begin{figure}
    \includegraphics[width=\textwidth]{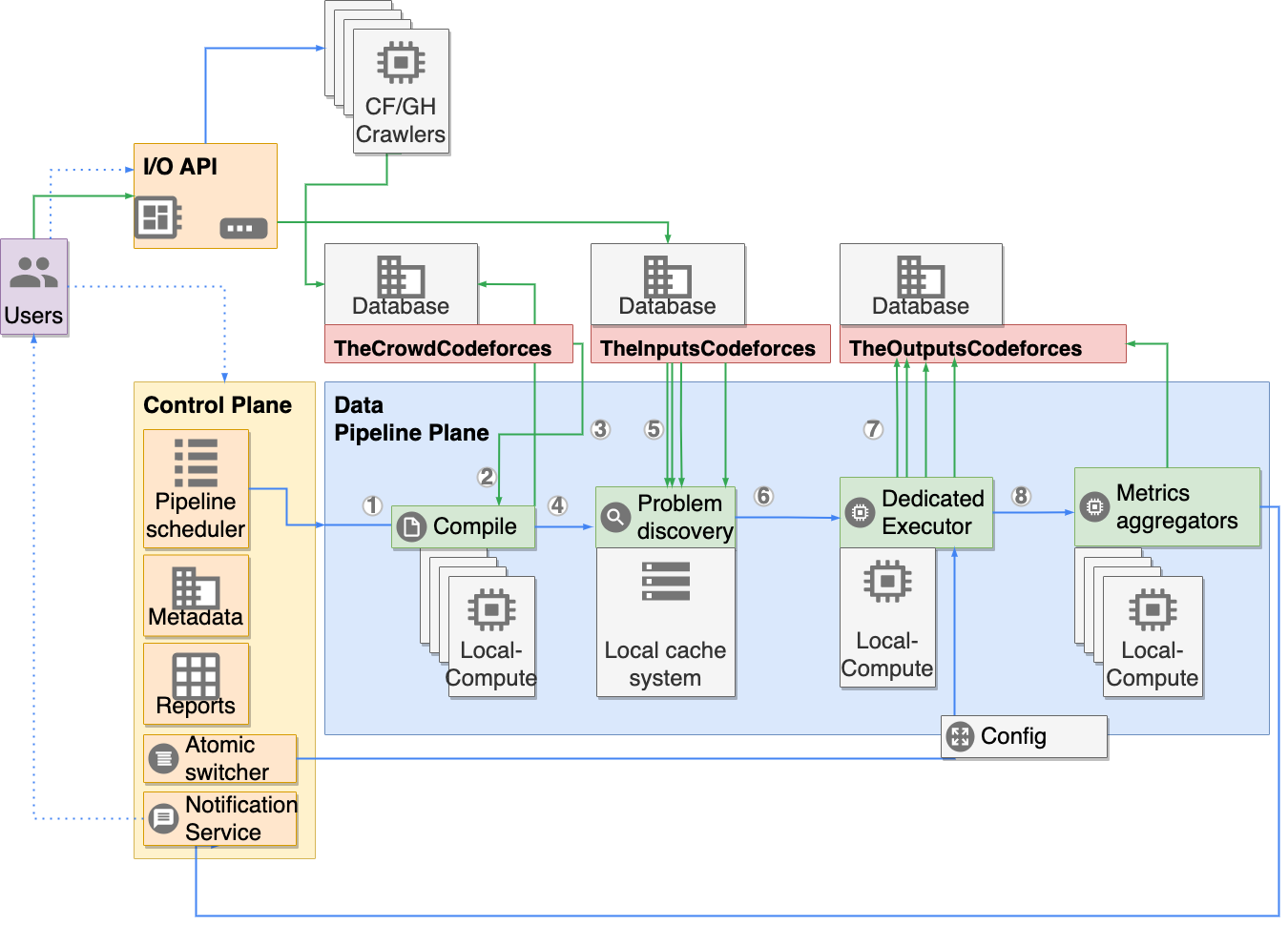}
    \caption{An overview of our data acquisition system, capable of generating a set of metrics from the binaries.}\label{data-acquisition-figure}
\end{figure}

\begin{figure}
    \centering
    \includegraphics[width=\textwidth]{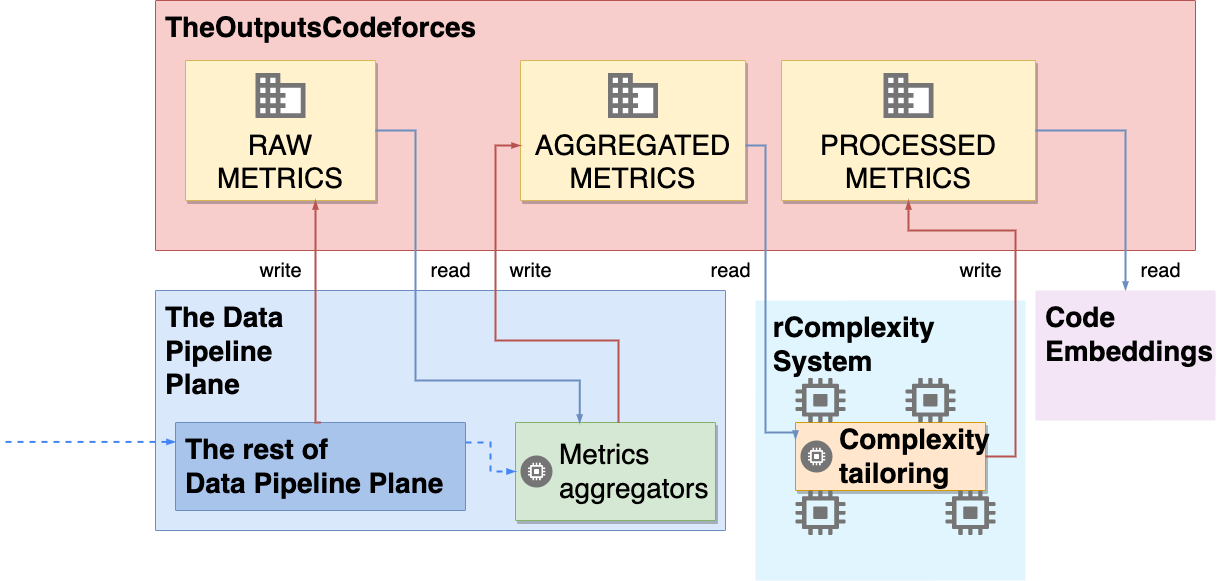}
    \caption{An overview of our data embedding system that converts the raw metrics into code embeddings, and how it interacts with other parts of the systems.}\label{emb-subsystem-figure}
\end{figure}

The \textit{system} we proposed to analyse the solutions to the problems contained in the dataset presented above is complex. It can be seen as two distinct entities: \textbf{the data acquisition subsystem} (Figure \ref{data-acquisition-figure}) and the \textbf{embeddings subsystem} (Figure \ref{emb-subsystem-figure}). The key steps involved in running the pipeline is illustrated by the labels in Figure \ref{data-acquisition-figure}. The Control Plane initiates the data pipeline in \textbf{1}, by providing the required arguments. Step \textbf{2} fetches all the matching solutions, to all problems, from TheCrawlCodeforces (the solutions dataset). It compiles the sources and store the results back in the database, as indicated in \textbf{3}. Next, in \textbf{4}, the data is passed, in form of a compiled executable, to the discovery service. Next, the synthetic inputs from TheInputsCodeforces\footnote{TheInputsCodeforces is a public dataset:\\ \url{https://github.com/raresraf/TheInputsCodeforces}} (input dataset) are being fetched, in \textbf{5}. Once this is complete, in step in \textbf{6}, the pipeline goes into a schedule state, where the execution on a dedicated executors is being prepared. In our research, to avoid variations in data due to difference in hardware, we have only used one executor, but this may have serious implication on the scalability. Once tasks have been scheduled and completed, the resulting profiling data is stored to the TheOutputsCodeforces (profiling dataset) in in \textbf{7}. At the end of the pipeline, metrics obtained from profiling are being aggregated in \textbf{8} and the pipeline is marked as succeeded.

After raw metrics with profiling data are being generated, the data embedding system (Figure \ref{emb-subsystem-figure}) aggregates them so that the complexity mapping can be applied. The embeddings system is responsible of building code embeddings from the available data provided by the Data Acquisition System. It is responsible of computing at scale dynamic code-embeddings based on r-Complexity with respect to the various metrics analyzed against the input dimensions.

Because the resulting embedding is somewhat dependent to the underlying architecture it has been profiled against, for best result, we evaluated all the solutions from our dataset on the same architecture. This doesn't make our solution architecture-dependent however, as we can re-process our metrics by launching the pipelines on any given architecture. It's also possible that similar are obtained even without performing this step when computing embeddings on new architecture, assuming that the two architectures are from the same family and have similar performances, or if there is a mapping function estimate regarding performance difference for the analyzed metrics.

\section{Dataset}\label{dataset}

\begin{figure}
    \centering
    \includegraphics[width=1\textwidth]{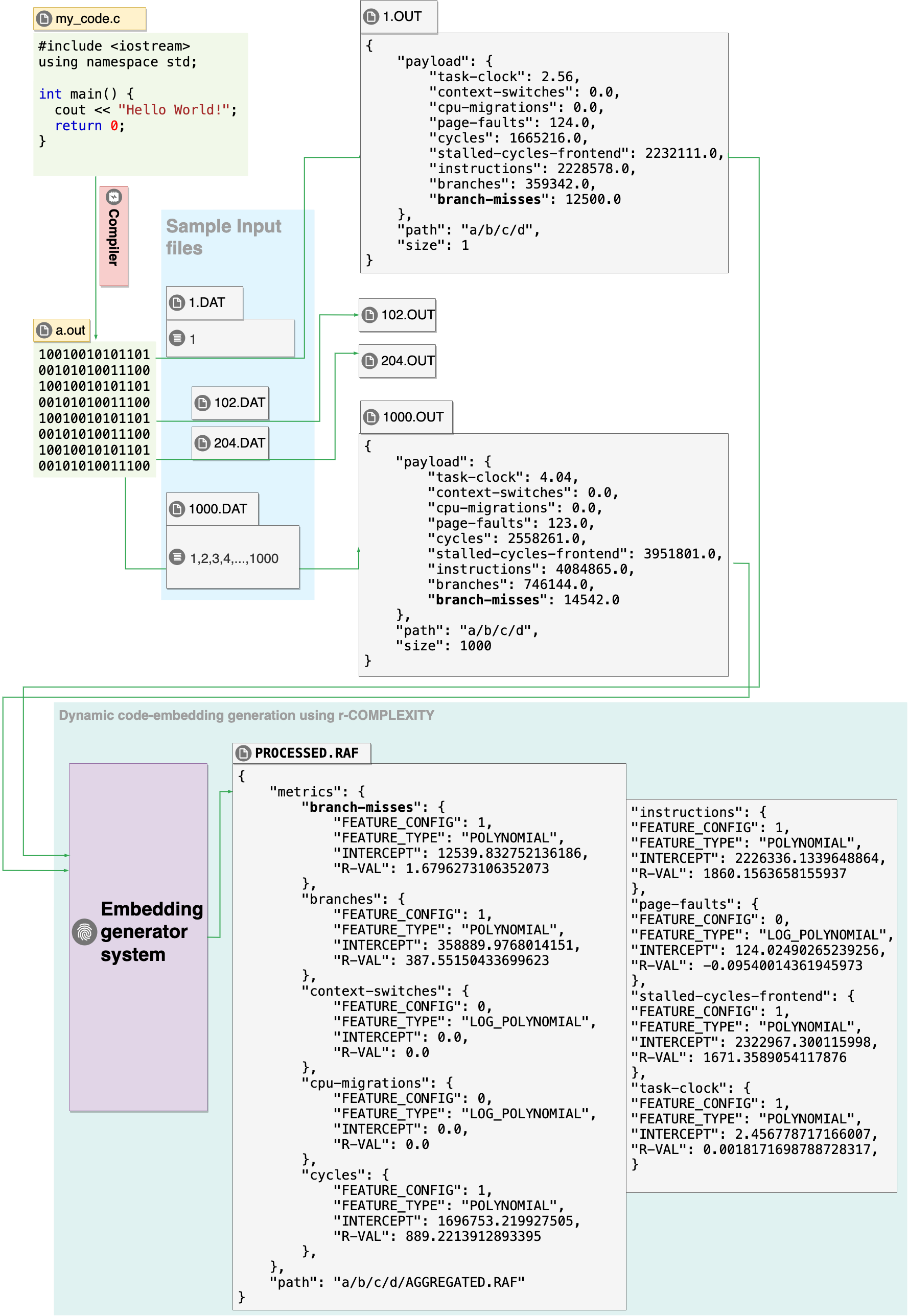}
    \caption{An in-depth view on the process of generating a code-embedding for a given algorithm, based on r-Complexity code-embeddings, using this system.}\label{dataset-figure}
\end{figure}

In this research, we built a automatic system that searches for open-source implementations for Codeforces technical challenges. We use the AlgoLabel~\cite{iacob2020algolabel} dataset in order to obtain the mapping between problems and algorithm labels. 

In order to achieve dynamic measurements out of which we can create the dynamic code embeddings, synthetic test inputs are required for each individual problem. The generation of such inputs is performed manually and usually requires some degree of reasoning, around choosing an appropriate dimension and input space for the problem, so that the samples will be relevant in computing the complexity. For instance, consider a problem, for which the number of required processing cycles to compute for the input $x$ is described by the function $c$:

\[
    c(x)= 
\begin{cases}
    10,              & x =  10000 \\
    1,              & x =  20000  \\
    100 \cdot {x}^2, & \text{otherwise }
\end{cases}
\]

For this problem, sampling the algorithm for any input is relevant in computing the associated r-Complexity, except of the results of analysing inputs $x=10000$ and $x=20000$, which are outliers. In practice, these outliers exists, and usually correspond to particular cases, when solving a difficult problem is trivial; for example, the problem of determining whether a number is prime or not is trivial for all even numbers, but has a non-constant complexity for arbitrary odd numbers. Hence, in order to provide relevant estimates of the r-Complexity functions for a given problem, the step of creating synthetic inputs requires some degree of judgement, to ensure that the inputs are representative for the difficulty of the problem.

This generation can be toilsome for a large dataset. In our research, we picked $50$ problems,
based on the number of available
sources. For every problem, we evaluate each compiled binary against a set of $50$ inputs. The trade-off here is the following: the more support points, the better the estimate for the complexity function becomes, but this comes at the cost of slowing down the pipeline. We validate empirically that this number of inputs may provide sufficient data points for inferring the program complexity.


Then, while executing the binaries on the previously generated inputs, the pipeline measures several statistics. These measurements are stored, obtaining the profiling files for each execution. Consequently, there are over one million such files in the resulting dataset (referenced as TheOutputsCodeforces dataset).

Next we create an embedding, using the techniques described in Section \ref{embedding-design}. Thus, each program is represented \textit{as a $36$-long vector}, built using approximations of the r-Theta complexity against the inputs, for each individual metric.
We depict this process in Figure \ref{dataset-figure}.

\section{Results}\label{results}
At this stage, we approach the classification problem of solution labelling with standard machine learning approaches. We study two scenarios: classifying each problem in a binary form, such as distributing each solution into belongs to category X or not, and performing multi-label classification, aiming to find all the classes a solution belongs to.

Three methods yield the best results: \textbf{decision tree classifier}, \textbf{random forest classifier} and \textbf{XGBoost}. We analyze the performance of two models based on multiple metrics: \textbf{accuracy}, \textbf{precision}, \textbf{recall} and \textbf{f1-score} on 5949 inputs, using a 66/34 split for the training/testing dataset. For all the analyzed methods, we perform the same split, simulating a similar training environment for all analyzed models.

In this research, several custom neural networks and deep neural networks, as well as convolutional networks, were trained, on the same datasets, but the overall performance was rather unsatisfactory. We could not maximise the performance of either to be close to the performance of a simple decision tree classifier. Going forward, we present the results of tree-based classifiers.

All models have been implemented in Python and published open-source\footnote{\url{https://www.github.com/raresraf/AlgoRAF}}. We use pandas to operate datasets, sklearn to compute metrics and for ensemble models, xgboost library for an efficient implementation of XGBoost trees and Tensorflow for experimenting with neural networks.

\subsection{Binary Classification}

\begin{figure}
    \includegraphics[width=\linewidth]{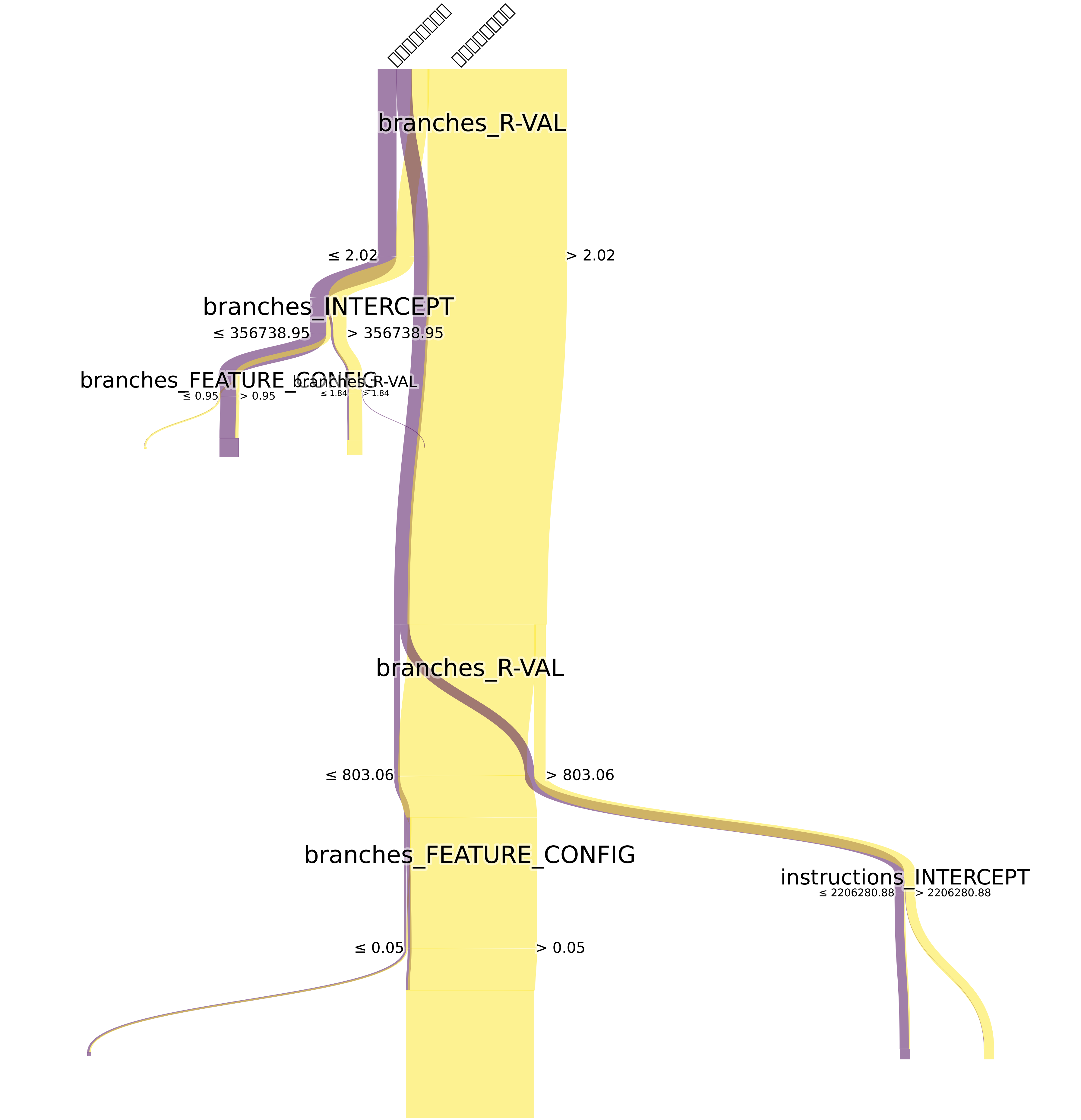}
    \caption{The decision tree classifier that has achieved over $96\%$ accuracy on the task of labelling algorithmic challenges with the math/non-math label against our testing dataset, while trained on dataset we prepared, with open-source Codeforces submissions. The figure contains the tree plotted with a maximum depth of three. The total depth of our decision trees was between twelve and sixteen. Purple are branches that evaluate to math class, while in yellow are plotted branches that evaluate to non-math class. }
\end{figure}

We perform binary classification for math/non-math problems. The dataset is unbalanced, with $4937$ problems not being labeled as math and $1012$ for problems that were classified as math-related. The aim was to keep the ratio from the original dataset in both testing and training dataset. The testing evaluation results are captured in~\textbf{Table 1}.

The accuracy is similar for both methods: \textbf{96\%} for the decision tree classifier and \textbf{97\%} for the random forest classifier. Overall, only the recall is smaller for the random forest classifier, while all the other metrics, including the f1-score is slightly higher for this classifier.

A visualisation of the decision tree model is available in Figure 4. High-quality renderings are available\footnote{\url{https://www.github.com/raresraf/AlgoRAF/tree/master/viz}}. 

\begin{table}
\centering
\caption{The metrics obtained for the \textbf{decision tree classifier} (DT) and \textbf{random forest classifier} (RF) evaluation against testing dataset.}
\begin{tabular}{|l|l|l|l|r|}
\hline
\textbf{ \ \ \ \ \ \ \ \  \ Metric}         & \textbf{precision} & \textbf{recall} & \textbf{f1-score} & \textbf{support} \\ 
   \textbf{Class}      & (DT $|$ RF) & (DT $|$ RF) & (DT $|$ RF) & \\ 

\hline
\textbf{non-math} & \textbf{.98} $|$ .97      & .98 $|$ \textbf{.99}  & \textbf{.98} $|$ \textbf{.98}    & 1663    \\ \hline
\textbf{math}     & .89 $|$ \textbf{.96}     & \textbf{.87} $|$ .83   & .88 $|$ \textbf{.89}    & 301     \\ \hline
\end{tabular}
\end{table}

\subsection{Multi-Label Classification}

For Multi-Label Classification, XGBoost yield the best results. In a similar setup, we benchmark the performances of a XGBoost classifier against the problem of algorithm classification. In our research, this method exhibits the best performance, in terms of \textit{precision}, \textit{recall} and \textit{F1-score} for the analyzed scenarios. Detailed results are provided in \textbf{Table 2}.

\setlength{\tabcolsep}{6pt}
\begin{table}[H]
\label{table:classification:report}
\centering
\caption{A classification report ran against the testing dataset, for the problem of algorithm classification, using the \textbf{XGBoost} classifier, a Decision Tree Classifier (DT) and using the random forest classifier technique (RF). In \textbf{bold}, we capture the results that are better than the previous two models, decision tree and random forest classifiers.}
\begin{tabular}{l | c | c | c | r}
\textbf{Class} & \textbf{Precision} & \textbf{Recall} & \textbf{F-score} & \textbf{Support}\\
\hdashline 
   &    XGBoost   &   XGBoost   &   XGBoost  &     \\
   &    DT $|$ RF   &   DT $|$ RF   &   DT $|$ RF  &     \\
\midrule
strings & \textbf{.94} & \textbf{.90} & \textbf{.92} & 756\\
 &    .86 $|$ .93   &   .87 $|$ .84   &   .87 $|$ .88   &     \\
\hdashline 
implementation & .94 & \textbf{.98} & \textbf{.96} & 1387\\
 &    .95 $|$ .93   &   .93 $|$ .97   &   .94 $|$ .95   &    \\
\hdashline 
greedy & \textbf{.92} & .77 & \textbf{.84} & 523\\
 &    .78 $|$ .91   &   .79 $|$ .65   &   .78 $|$ .76   &     \\
\hdashline 
brute force & .98 & .77 & \textbf{.86} & 311\\
 &    .79 $|$ 1.0   &   .77 $|$ .66   &   .78 $|$ .79   &     \\
\hdashline 
dp & \textbf{.87} & .74 & .80 & 35\\
 &    .83 $|$ .87   &   .71 $|$ .77   &   .77 $|$ .82   &      \\
\hdashline 
divide and conquer & \textbf{1.0} & .68 & .81 & 31\\
 &    .91 $|$ 1.0   &   .65 $|$ .71   &   .75 $|$ .83   &      \\
\hdashline 
graphs & .91 & \textbf{.88} & \textbf{.90} & 83\\
 &    .92 $|$ .84   &   .82 $|$ .83   &   .87 $|$ .84   &      \\
\hdashline 
binary search & \textbf{1.0} & .68 & .81 & 31\\
 &    .91 $|$ 1.0   &   .65 $|$ .71   &   .75 $|$ .83   &      \\
\hdashline 
math & \textbf{.97} & \textbf{.91} & \textbf{.94} & 301\\
 &    .91 $|$ .95   &   .85 $|$ .81   &   .88 $|$ .88   &     \\
\hdashline 
sortings & .95 & .61 & \textbf{.74} & 176\\
 &    .66 $|$ .96   &   .66 $|$ .42   &   .66 $|$ .58   &     \\
\hdashline 
shortest paths & .91 & \textbf{.88} & \textbf{.90} & 83\\
  &    .92 $|$ .84   &   .82 $|$ .83   &   .87 $|$ .84   &      \\
\midrule
micro avg & \textbf{.94} & \textbf{.88} & \textbf{.91} & 3717\\
    &    .87 $|$ .93   &   .85 $|$ .82   &   .86 $|$ .87  &    \\
\hdashline 
macro avg & \textbf{.94} & \textbf{.80} & \textbf{.86} & 3717\\
    &    .86 $|$ .93   &   .77 $|$ .74   &   .81 $|$ .82  &    \\
\hdashline 
weighted avg & \textbf{.94} & \textbf{.88} & \textbf{.91} & 3717\\
    &    .87 $|$ .93   &   .85 $|$ .82   &   .86 $|$ .86  &    \\
\hdashline 
samples avg & \textbf{.94} & \textbf{.91} & \textbf{.91} & 3717\\
    &    .89 $|$ .91   &   .88 $|$ .86   &   .88 $|$ .87  &    \\
\end{tabular}
\end{table}

The model scores really well in the precision of identifying a problem belonging to the class. Slightly worse, the recall is diminished, especially for classes where the number of support entries is low, such as \textit{divide and conquer} or \textit{binary search}. It's interesting that the model does match all the 30+ entries in the testing dataset where these two techniques are used, yet, a recall of just above two thirds is recorded. However, for large sets, such as strings or math classes, the model scores very high in both precision and recall, with sustained results of over $90\%$.

Our experiments also included the use of decision trees in junction with \textit{r-Complexity} embeddings, as well as random forests classifiers. Overall, the results obtaining this ensemble model are better than by using a single, simple, decision tree classifier, for almost all classes and all analyzed metrics. Between these two methods, it seems that the Random Forest Classifier is optimized for higher precision, with a slightly lower recall score. Just like XGBoost, the model does match all the entries for the \textit{divide and conquer} and \textit{binary search}, and even obtains a higher recall score for these two classes than the XGBoost classifier. However, the average of recall is $4\%$ smaller. 

\section{Conclusions and further research}\label{conclusions}

In this paper we present a way of building code embeddings based on Complexity related measurements and use these telemetry as insights for the problem of algorithm classification. The method showcased for transforming the source code to numerical embeddings is a helpful solution when needing to analyze the behaviour of algorithms. Using these embeddings that are based on r-Complexity~\cite{folea2021new}, we achieve high accuracy classification even when using simple decision tree-based classifiers for the problem of labelling of competitive programming challenges. 

This shows the potential of this embedding method, for building a more general understanding of the source code. The greater goal of our research is to build a generic model of mapping an algorithm to a code embedding (based on the r-Complexity code embeddings technique), that can later be used for more competitive and advanced programming challenges. We believe that these embeddings covers a good amount of information about the code, and can be used to further analysis, such as: \textit{plagiarism detection}, \textit{software optimisations and classification} or \textit{malware detection}. Tailoring the collected metrics for a certain algorithm can help reach this goal, by introducing other more relevant metrics for solving a certain task. 

Further work on the development of the TheInputsCodeforces database, in order to support more generators, that would produce more synthetic inputs. This might enhance the generality of our dataset and is likely to make our models more relevant to a wider set of problems.

%
%
%
\bibliographystyle{splncs04}
%
\bibliography{references}

@preamble{"\newcommand{\SortNoop}[1]{}"}

@article{alon2019code2vec,
  title={code2vec: Learning distributed representations of code},
  author={Alon, Uri and Zilberstein, Meital and Levy, Omer and Yahav, Eran},
  journal={Proceedings of the ACM on Programming Languages},
  volume={3},
  number={POPL},
  pages={1--29},
  year={2019},
  publisher={ACM New York, NY, USA}
}

@inproceedings{Liger2020,
author = {Wang, Ke and Su, Zhendong},
title = {Blended, Precise Semantic Program Embeddings},
year = {2020},
isbn = {9781450376136},
publisher = {Association for Computing Machinery},
address = {New York, NY, USA},
url = {https://doi.org/10.1145/3385412.3385999},
doi = {10.1145/3385412.3385999},
booktitle = {Proceedings of the 41st ACM SIGPLAN Conference on Programming Language Design and Implementation},
pages = {121–134},
numpages = {14},
location = {London, UK},
series = {PLDI 2020}
}

@inproceedings{koc2017learning,
  title={Learning a classifier for false positive error reports emitted by static code analysis tools},
  author={Koc, Ugur and Saadatpanah, Parsa and Foster, Jeffrey S and Porter, Adam A},
  booktitle={Proceedings of the 1st ACM SIGPLAN International Workshop on Machine Learning and Programming Languages},
  pages={35--42},
  year={2017}
}

@inproceedings{DBLP:conf/iclr/WangSS18,
  author    = {Ke Wang and
               Rishabh Singh and
               Zhendong Su},
  title     = {Dynamic Neural Program Embeddings for Program Repair},
  booktitle = {6th International Conference on Learning Representations, {ICLR} 2018,
               Vancouver, BC, Canada, April 30 - May 3, 2018, Conference Track Proceedings},
  publisher = {OpenReview.net},
  year      = {2018},
  url       = {https://openreview.net/forum?id=BJuWrGW0Z},
  bibsource = {dblp computer science bibliography, https://dblp.org}
}

@article{dypro,
  author    = {Ke Wang},
  title     = {Learning Scalable and Precise Representation of Program Semantics},
  journal   = {CoRR},
  volume    = {abs/1905.05251},
  year      = {2019},
  url       = {http://arxiv.org/abs/1905.05251},
  eprinttype = {arXiv},
  eprint    = {1905.05251},
  bibsource = {dblp computer science bibliography, https://dblp.org}
}

@inproceedings{neuralcodecomprehension,
author = {Ben-Nun, Tal and Jakobovits, Alice Shoshana and Hoefler, Torsten},
title = {Neural Code Comprehension: A Learnable Representation of Code Semantics},
year = {2018},
publisher = {Curran Associates Inc.},
address = {Red Hook, NY, USA},
booktitle = {Proceedings of the 32nd International Conference on Neural Information Processing Systems},
pages = {3589–3601},
numpages = {13},
location = {Montr\'{e}al, Canada},
series = {NIPS'18}
}

@article{DBLP:journals/corr/abs-1812-09652,
  author    = {Kimberly Redmond and
               Lannan Luo and
               Qiang Zeng},
  title     = {A Cross-Architecture Instruction Embedding Model for Natural Language
               Processing-Inspired Binary Code Analysis},
  journal   = {CoRR},
  volume    = {abs/1812.09652},
  year      = {2018},
  url       = {http://arxiv.org/abs/1812.09652},
  eprinttype = {arXiv},
  eprint    = {1812.09652},
  bibsource = {dblp computer science bibliography, https://dblp.org}
}

@article{DBLP:journals/corr/abs-1804-03635,
  author    = {Alexander Chistyakov and
               Ekaterina Lobacheva and
               Arseny Kuznetsov and
               Alexey Romanenko},
  title     = {Semantic embeddings for program behavior patterns},
  journal   = {CoRR},
  volume    = {abs/1804.03635},
  year      = {2018},
  url       = {http://arxiv.org/abs/1804.03635},
  eprinttype = {arXiv},
  eprint    = {1804.03635},
  bibsource = {dblp computer science bibliography, https://dblp.org}
}

@article{DBLP:journals/corr/abs-2006-12641,
  author    = {Luca Buratti and
               Saurabh Pujar and
               Mihaela A. Bornea and
               J. Scott McCarley and
               Yunhui Zheng and
               Gaetano Rossiello and
               Alessandro Morari and
               Jim Laredo and
               Veronika Thost and
               Yufan Zhuang and
               Giacomo Domeniconi},
  title     = {Exploring Software Naturalness through Neural Language Models},
  journal   = {CoRR},
  volume    = {abs/2006.12641},
  year      = {2020},
  url       = {https://arxiv.org/abs/2006.12641},
  eprinttype = {arXiv},
  eprint    = {2006.12641},
  bibsource = {dblp computer science bibliography, https://dblp.org}
}

@inproceedings{svyatkovskiy2021fast,
  title={Fast and memory-efficient neural code completion},
  author={Svyatkovskiy, Alexey and Lee, Sebastian and Hadjitofi, Anna and Riechert, Maik and Franco, Juliana Vicente and Allamanis, Miltiadis},
  booktitle={2021 IEEE/ACM 18th International Conference on Mining Software Repositories (MSR)},
  pages={329--340},
  year={2021},
  organization={IEEE}
}

@inproceedings{mikolov-word2vec-2018-advances,
    title = "Advances in Pre-Training Distributed Word Representations",
    author = "Mikolov, Tomas  and
      Grave, Edouard  and
      Bojanowski, Piotr  and
      Puhrsch, Christian  and
      Joulin, Armand",
    booktitle = "Proceedings of the Eleventh International Conference on Language Resources and Evaluation ({LREC} 2018)",
    month = may,
    year = "2018",
    address = "Miyazaki, Japan",
    publisher = "European Language Resources Association (ELRA)",
    url = "https://aclanthology.org/L18-1008",
}

@inproceedings{folea2021new,
  title={A new metric for evaluating the performance and complexity of computer programs: A new approach to the traditional ways of measuring the complexity of algorithms and estimating running times},
  author={Folea, Rares and Slusanschi, Emil-Ioan},
  booktitle={2021 23rd International Conference on Control Systems and Computer Science (CSCS)},
  pages={157--164},
  year={2021},
  organization={IEEE}
}

@inproceedings{chen2016xgboost,
  title={Xgboost: A scalable tree boosting system},
  author={Chen, Tianqi and Guestrin, Carlos},
  booktitle={Proceedings of the 22nd acm sigkdd international conference on knowledge discovery and data mining},
  pages={785--794},
  year={2016}
}

@article{calotoiu2018automatic,
  title={Automatic empirical performance modeling of parallel programs},
  author={Calotoiu, Alexandru},
  year={2018},
  publisher={Technische Universit{\"a}t}
}

@article{iacob2020algolabel,
  title={AlgoLabel: A Large Dataset for Multi-Label Classification of Algorithmic Challenges},
  author={Iacob, Radu Cristian Alexandru and Monea, Vlad Cristian and R{\u{a}}dulescu, Dan and Ceap{\u{a}}, Andrei-Florin and Rebedea, Traian and Tr{\u{a}}usan-Matu, Stefan},
  journal={Mathematics},
  volume={8},
  number={11},
  pages={1995},
  year={2020},
  publisher={MDPI}
}

@inproceedings{yousefi2018learning,
  title={Learning latent byte-level feature representation for malware detection},
  author={Yousefi-Azar, Mahmood and Hamey, Len and Varadharajan, Vijay and Chen, Shiping},
  booktitle={Neural Information Processing: 25th International Conference, ICONIP 2018, Siem Reap, Cambodia, December 13-16, 2018, Proceedings, Part IV 25},
  pages={568--578},
  year={2018},
  organization={Springer}
}


\end{document}